\documentclass[runningheads]{llncs}
\usepackage{graphicx}
\usepackage{booktabs}
\usepackage{url}
\usepackage{hyperref}
\usepackage{amsmath}
\usepackage{soul}
\usepackage{multirow}

\begin{document}
%
\title{Automated Pulmonary Embolism Detection from CTPA Images Using an End-to-End Convolutional Neural Network}
\titlerunning{PE Detection Using End-to-End CNN}
%
\author{
Yi Lin\inst{1}
\and
Jianchao Su\inst{1}
\and
Xiang Wang\inst{2}
\and
Xiang Li\inst{2}
\and
Jingen Liu\inst{3}
\and
Kwang-Ting Cheng\inst{4}
\and
Xin Yang*\inst{1}
}
\authorrunning{Y. Lin et al.}
%
\institute{Huazhong University of Science and Technology, China
\and
The Central Hospital of Wuhan, China
\and
JD AI Research, Mountain View, USA
\and
Hong Kong University of Science and Technology, Hong Kong\\
	*corresponding author~\email{xinyang2014@hust.edu.cn}
}
\maketitle              
\begin{abstract}
Automated methods for detecting pulmonary embolisms (PEs) on CT pulmonary angiography (CTPA) images are of high demand. Existing methods typically employ separate steps for PE candidate detection and false positive removal, without considering the ability of the other step. As a result, most existing methods usually suffer from a high false positive rate in order to achieve an acceptable sensitivity. This study presents an end-to-end trainable convolutional neural network (CNN) where the two steps are optimized jointly. The proposed CNN consists of three concatenated subnets: 1) a novel 3D candidate proposal network for detecting cubes containing suspected PEs, 2) a 3D spatial transformation subnet for generating fixed-sized vessel-aligned image representation for candidates, and 3) a 2D classification network which takes the three cross-sections of the transformed cubes as input and eliminates false positives. We have evaluated our approach using the 20 CTPA test dataset from the PE challenge, achieving a sensitivity of 78.9\%, 80.7\% and 80.7\% at 2 false positives per volume at 0mm, 2mm and 5mm localization error, which is superior to the state-of-the-art methods. We have further evaluated our system on our own dataset consisting of 129 CTPA data with a total of 269 emboli. Our system achieves a sensitivity of 63.2\%, 78.9\% and 86.8\% at 2 false positives per volume at 0mm, 2mm and 5mm localization error.

\end{abstract}
\section{Introduction}
Pulmonary Embolism (PE) is a thrombus that obstructs central, lobar, segmental, or subsegmental pulmonary arteries when it travels from the heart to the lungs. The morality rate of untreated PE is around 30\%; however, early detection and treatment of PE could effectively decrease the morality rate to as low as 2\% \cite{tajbakhsh2015computer}. Computed tomography pulmonary angiography (CTPA), in which PE appears as a filling defect (i.e. a dark region surrounded by the bright vessel lumen), is the primary means for diagnosing PE in today's clinical practice. However, manually interpreting a CTPA volume demands a radiologist to carefully trace each pulmonary artery across 300-500 slices for any suspected PEs, which is time consuming and could suffer from a large inter-/intra-observer variety due to the different radiologists' experience, attention span and eye fatigue. 

Automated detection of PE is of high demand for improving the accuracy and efficiency of PE detection and diagnosis. Existing methods typically consists of two separate steps: 1) generating a set of PE candidates based on voxel-level features; and 2) extracting region-level features of PE candidates and eliminating false positives (FPs) based on a classifier. For example, in \cite{masutani2002computerized} Masutani et al. extracted handcrafted features based on CT values, local contrast and the second derivatives of voxels for candidate detection. The volume, effective length and mean local contrast of grouped voxels were then extracted from the detected candidates for false positive removal. In \cite{bouma2009automatic}, Bouma et al. used similar voxel-level features as \cite{masutani2002computerized} for candidate proposal and developed new region-level features based on the isophote curvature and circularity of the bright lumen for FP removal. However, due to the limited representation ability of these handcrafted features, conventional methods often suffer from a high number of false positives in order to achieve an acceptable sensitivity. Nima et al. \cite{tajbakhsh2015computer} investigated the feasibility of CNN features for eliminating false positives in the task of PE detection. A novel vessel-aligned multi-planar image representation based on CNN was developed in \cite{tajbakhsh2015computer} and demonstrated to be highly effective, efficient and robust for distinguishing true PEs from false positives. In this study, we also adopt such vessel-aligned image representation for its good robustness and distinctiveness. While different from \cite{tajbakhsh2015computer} which utilized handcrafted feature based method \cite{liang2007computer} for detecting candidate PEs and a separate CNN for FP removal, this work for the first time integrates all steps including PE candidate proposal, the generation of vessel-aligned image representation and FP removal into an end-to-end trainable CNN for a jointly optimized PE detector.

As shown in Fig. \ref{fig1}, our PE detection network is a cascade of three subnets: candidate proposal subnet, spatial transformation network, and false positive removal network. The first subnet performs PE candidate proposal via a 3D fully convolutional network (FCN) to extract 3D feature hierarchies, followed by two 3D convolutional layers to generate candidate cubes containing PEs.  The second subnet aims to transform the candidate cubes so as to align the suspected embolus with the orientation of the affected vessel segment. We implement the goal using two parametric-free layers, i.e. a spatial transformation layer and a 3D region-of-interest (ROI) pooling layer, which are derivable and can facilitate error back-propagation in end-to-end training. The third subnet is a 2D classification CNN which takes three cross-sections of the transformed candidate cube as input and outputs its probability of being a PE. Our method is very efficient since convolutional features are shared among the proposal, spatial transformation and classification pipelines. We have evaluated our approach using the entire 20 CTPA test dataset from the PE challenge, achieving a sensitivity of 78.9\%, 80.7\% and 80.7\% at 2 false positives per volume at 0mm, 2mm and 5mm localization error. This performance is superior to the winning system in the literature, which achieves a sensitivity of 60.5\%, 66.4\% and 75.8\% at the same level of false positives. We have further evaluated our system on our own dataset consisting of 129 CTPA data. Our system achieves a sensitivity of 63.2\%, 78.9\% and 86.8\% at 2 false positives per volume at 0mm, 2mm and 5mm localization error.

To summarize, the main contributions of our paper are:
\begin{itemize}
	\item[$\bullet$] an end-to-end PE detection network with combined candidate proposal, spatial transformation for vessel-aligned image representation, and classification stages that detects arbitrary size embolisms;
	\item[$\bullet$] a derivable and parametric-free implementation of vessel-aligned representation using a 3D spatial transformation subnet;
	\item[$\bullet$] extensive evaluations on two diverse datasets consists of totally 149 patients	data that demonstrate the general applicability and superior performance of our model to the state-of-the-art methods.
\end{itemize}

\section{Method}
\begin{figure}[!t]
	\centering
	\includegraphics[width=\textwidth]{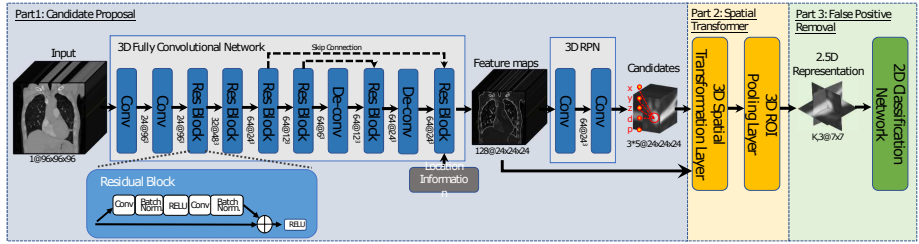}
	\caption{Framework of our end-to-end PE detection network.} \label{fig1}
	\vspace{-0.5cm}
\end{figure}
Fig. \ref{fig1} illustrates the framework of our PE detection network which consists of three subnets: a 3D candidate proposal subnet, a derivable 3D spatial transformation subnet, and a false positive removal subnet.

\subsection{Candidate Proposal Subnet} 
\textbf{3D FCN for feature extraction.} Our proposal subnet extracts features using a 3D FCN that employs an autoencoder network architecture with skip connections. Specifically, the encoder starts with two 3D convolution layers, followed by a max-pooling layer. Then, four residual blocks \cite{he2016deep} are applied to encode hierarchical feature maps. In the decoder,  the feature maps are up-sampled by two de-convolution layers and two residual blocks. Skip connections are utilized to connect the last two residual blocks in the encoder and the corresponding two residual blocks in the decoder. The size of a 3D pulmonary CTPA volume is typically very large, e.g. $512\times512\times400$. Directly inputting the entire 3D volume into the network could involve extremely high memory cost. To meet the constraint of GPU memory, we divide the entire volume into overlapping cubes size of $96\times96\times96$ and input each cube into the network, rather than the entire 3D volume. We also combine location information with the FCN feature maps as we believe location is an important indicator for identifying PEs which usually reside at some unique locations, i.e. bifurcations of the main pulmonary artery or lobar branches \cite{araoz2007pulmonary}. To this end, we form a feature map with the same size as FCN feature maps. Each voxel of the location feature map is a 3-dimensional vector indicating the $x$, $y$ and $z$ coordinates in the entire 3D volume. We concatenate this 3-channel feature map with the FCN feature map.

\textbf{3D region proposal.} The size of PEs could vary significantly depending on its locations in pulmonary arteries. Inspired by Faster R-CNN \cite{ren2015fasterarxiv}, we incorporate anchor cubes into a region proposal subnet to facilitate accurate detection of variable size proposals. Specifically, the anchor cues are pre-defined multi-scale 3D windows centered at every voxel location of the feature map. Each voxel location specifies $N=3$ anchor cubes, each at a different fixed scale $s$ (i.e. $s = 10mm$, $30mm$ and $60mm$ respectively) to ensure that the proposal prediction is scale invariant. For each anchor cube of a scale $s$, we design five detectors to regress five values ($\Delta x_s, \Delta y_s, \Delta z_s, \Delta d_s, p_s$) which are offsets in locations ($\Delta x_s, \Delta y_s, \Delta z_s$) and size ($\Delta d_s$) of the proposal cube with respect to the anchor cubes, as well as the probability ($p$) of the proposal containing a PE. To this end, we first apply a 3D convolution layer with 64 kernels size of $1\times1\times1$ to the feature map, then we apply another 3D convolution layer with $5×N$ kernels size of $1\times1\times1$ to output a feature map, each voxel of which is a $5×N$-dimensional vector indicating ($\Delta x_s, \Delta y_s, \Delta z_s, \Delta d_s, p_s$), $s\in\{1, ..., N\}$. 

\textbf{Training.} To train the candidate proposal subnet, we assign a binary class label to each anchor. If an anchor overlaps with some ground-truth with Intersection-over-Union (IoU) greater than 0.5, we label it as positive. If the anchor has an IoU overlap smaller than 0.02 with all ground-truth, we label it as negative. Anchors that are neither positive nor negative are excluded for training. The objective function of the candidate proposal subnet is defined as:
\begin{equation}
\label{eq1}
L(\{p_i\},\{\boldsymbol{t_i}\})=\frac{1}{N_{cls}}\sum_iL_{cls}(p_i,p^*_i)+\lambda\frac{1}{N_{reg}}\sum_i p^*_i L_{reg} (\boldsymbol{t_i},\boldsymbol{t^*_i})
\end{equation}
which consists of two losses:  the classification loss $L_{cls}$ is the binary cross entropy loss, and the regression loss $L_{reg}$ is the smooth $L_1$ loss. This two terms are normalized by $N_{cls}$ and $N_{reg}$, and weighted by $\lambda$. Note that $L_{reg}$ only applies to the positive anchor. In Eq. \ref{eq1}, $i$ denotes the $i$th anchor in a mini-batch, $p_i$ and $p^*_i$ denote the predicted probability of being PE and the ground-truth label ($p^* = \{0, 1\}$). $\boldsymbol{t_i}$ and $\boldsymbol{t^*_i}$ denote the predicted position and ground-truth position associated with a positive anchor, which consists of 4 parameters:
\begin{equation}
\begin{aligned}
&\Delta x=(x-x_a)/d_a,\	\Delta y=(y-y_a)/d_a,\\
&\Delta z=(z-z_a)/d_a,\	\Delta d=log(d/d_a)\\
\end{aligned}
\end{equation}
where ($x,y,z,d$) are predicted or ground-truth cube's center coordinates and its side length, and ($x_a,y_a,z_a,d_a$) are for anchor cube.

We use online hard sample mining to balance hard and easy samples in the training phase. In each mini-batch we randomly select $M$ negative samples, sort them in a descending order based on their classification score. The top $k$ samples are chosen as hard samples and contribute to the calculation of the loss function, the rest are abandoned by setting its loss to 0.
\begin{figure}[!t]
	\centering
	\includegraphics[width=\textwidth]{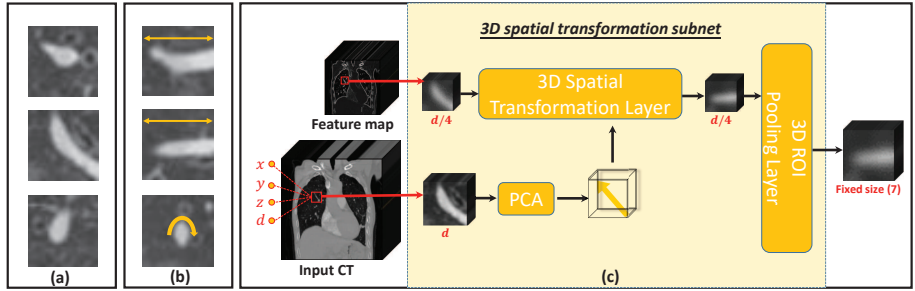}
	\caption{(a)-(b) cross-section views of a proposal cube before and after being processed by the 3D spatial transformation subnet. The yellow arrows denote $\boldsymbol{v_1}$, $\boldsymbol{v_2}$ and $\boldsymbol{v_3}$ the affected vessel. (c) Vessel-aligned fixed-sized proposal cube generation.} \label{fig2}
\end{figure}
\subsection{3D Spatial Transformation Subnet}
The proposal network could generate many false positives in order to achieve a satisfactory sensitivity. To exclude false positives, an intuitive solution is to apply an addition classification CNN which takes the proposal cubes as input and distinguish between true PE cubes and false positives. The problem, however, is that the appearance variations of all possible PEs is very large in practice. Training a classifier using a limited number of PE samples could yield severe overfitting and in turn exhibit poor performance for unseen testing cases. To alleviate this problem, Nima et al. \cite{tajbakhsh2015computer} proposed a vessel-aligned image representation, which aligns each candidate proposal to the orientation of the affected vessel to reduce the appearance variations of PEs in the three cross-section slices. Inspired by their method, in this work we develop a 3D spatial transformation subnet which generates vessel-aligned fixed-sized cubes from arbitrary-sized candidate proposals (see Fig. \ref{fig2}). The vessel-aligned 2.5D representation (as shown in Fig. \ref{fig2}(b)) of the candidate is then derived by extracting the axial, sagittal and coronal slices from the vessel-aligned cube. 

To implement our 3D spatial transformation subnet, we crop two proposal cubes (i.e. $C_{ori}$ and $C_{feat}$) from both the original CTPA volume and the corresponding feature maps with stride of four, respectively. We compute the vessel orientation from $C_{ori}$ by first segmenting vessels via simple intensity thresholding. Then we apply principle component analysis (PCA) to the segmented vessels to obtain three eigen vectors ($\boldsymbol{v_1}, \boldsymbol{v_2}, \boldsymbol{v_3}$) and their corresponding eigen value ($\lambda_1 \ge \lambda_2 \ge \lambda_3$). The first eigen vector $\boldsymbol{v_1}$ represents the direction in which the artery elongates, while $\boldsymbol{v_2}$ and $\boldsymbol{v_3}$ represent two orthogonal directions in the plane vertical to $\boldsymbol{v_1}$. We align $C_{feat}$ to vessel's orientation via the 3D affine transformation \cite{jaderberg2015spatial}, according to Eq. \ref{eq3}:
\begin{equation}
\label{eq3}
{\left
	[\begin{array}{c}
	x^s \\
	y^s \\
	z^s
	\end{array}
	\right]} = A_\theta
{\left[
	\begin{array}{c}
	x^t\\
	y^t\\
	z^t\\
	1
	\end{array}
	\right]}=
{\left[
	\begin{array}{cccc}
	s_x\boldsymbol{e}^T_1\boldsymbol{v}_1 & s_y\boldsymbol{e}^T_1\boldsymbol{v}_2 & s_z\boldsymbol{e}^T_1\boldsymbol{v}_3 & t_x \\
	s_x\boldsymbol{e}^T_2\boldsymbol{v}_1 & s_y\boldsymbol{e}^T_2\boldsymbol{v}_2 & s_z\boldsymbol{e}^T_2\boldsymbol{v}_3 & t_y \\
	s_x\boldsymbol{e}^T_3\boldsymbol{v}_1 & s_y\boldsymbol{e}^T_3\boldsymbol{v}_2 & s_z\boldsymbol{e}^T_3\boldsymbol{v}_3 & t_z 
	\end{array}
	\right]}
{\left[
	\begin{array}{c}
	x^t\\
	y^t\\
	z^t\\
	1
	\end{array}	
	\right]}
\end{equation}
where ($x^t,y^t,z^t$) and ($x^s,y^s,z^s$) are the normalized target coordinates of the regular grid in the transformed cube and the normalized source coordinates in the feature map. $A_\theta$ is the affine transformation matrix, where $t_x, t_y, t_z$ represent the offsets of the candidate center to the CT center, $s_x, s_y, s_z$ represent the scaling ratio between the candidate's size and the CT's size, $\boldsymbol{e}_1, \boldsymbol{e}_2, \boldsymbol{e}_3$ forms a unit matrix. We input vessel-aligned feature maps $C_{feat}$ to a 3D ROI pooling layer to extract fixed-sized feature maps (i.e. $7\times 7\times 7$ ) for FP removal. Fig. \ref{fig2}(c) illustrates the pipeline of 3D spatial transformation subnet.

\subsection{False Positive Removal Subnet}
We extract three cross sections to form a 2.5D representation and feed it to a simple classification subnet consisting of two fully connected layers for FP removal. In training, a sample is labeled as positive if its central point is within an embolus. Otherwise, it is considered as negative. To avoid any bias to negative samples, we randomly select 128 training samples in an iteration, where the radio of positive to negative samples is up to 1:3. If there are insufficient positive samples in an iteration, we pad negative ones in the mini-batch. 

\section{Experiments}
\subsection{Datasets}
We evaluated our method using two datasets: 1) the test set of the PE challenge \cite{pechallenge} consisting of 20 patients CTPA scans, and 2) the PE129 dataset consisting 129 CTPA scans with a total of 269 embolisms. 99 scans of the PE129 are collected in a local hospital and 30 of them are from \cite{masoudi2018new}.


\subsection{Implementation Details}
We pre-process the CTPA data before inputting it to our network. First, we segment the lung region to exclude the background tissues based on \cite{liao2019evaluate}. Then, we re-sample the data to achieve an isotropic resolution ($1mm\times1mm\times1mm$). After that, we adjust the contrast by clipping the data into [-1200, 600] HU and linearly transforming it into [-1, 1].

To train the candidate proposal subnet, we augment the dataset by randomly flipping, rotating ($0\sim 180^{\circ}$) and scaling ($0.75X\sim 1.25X$) each original sample. The 3D FCN was pretrained on a large public dataset LUNA16 for nodule detection. We train the candidate proposal subnet in 100 epochs using SGD optimization with the learning rate 0.001, momentum 0.9, weight decay $10^{-4}$.


\begin{figure}[!t]
	\centering
	\includegraphics[width=\textwidth]{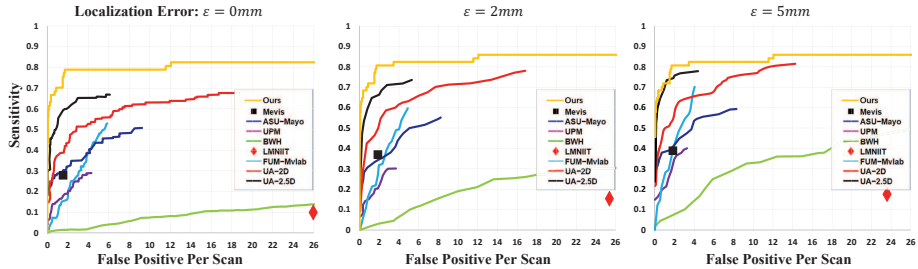}
	\caption{Comparison with the state-of-the-art methods on the PE challenge dataset. We plot the FROC curve per scan at 0mm, 2mm and 5mm localization error.
	} \label{fig3}
\end{figure}

\subsection{Comparison with Other Methods} 
We compared our PE detection network with the methods \cite{pechallenge} on the PE challenge test set. As the ground truth labels are not available on the website, we asked a radiologist with over 10 years' reading experience to manually delineate PEs in each test scan. The manual annotations were further validated by a 2nd observer. We will also release our manual annotations for a fair comparison in the future studies. The evaluation metrics is FROC curve per CTPA scan. A detection is counted as positive if it locates within a certain range (i.e. 0mm, 2mm and 5mm) to an embolus manual mask. Fig. \ref{fig3} shows that our method achieves a sensitivity of 78.9\%, 80.7\% and 80.7\% at 2FPs/scan when the localization error is 0mm, 2mm and 5mm, outperforming the winning system UA-2.5D which achieves 60.8\%, 66.4\% and 75.8\% at 2FPs/scan at 0mm, 2mm and 5mm localization error.
\begin{table}[!t]
	\centering
	\caption{Impact of the three subnets in our PE detection candidate network. Numbers indicate sensitivities at 2 FPs/scan at 0mm localization error.}
	\label{tab1}
	\setlength{\tabcolsep}{7mm}{
	\begin{tabular}{l|c|c}
		\toprule
		Method & PE challenge & PE129\\
		\hline
		$\mathrm{S_1}$ &  71.9\% & 42.1\%\\
		$\mathrm{S_1 + S_3}$ & 73.7\% & 44.7\%\\
		$\mathrm{S_1 + S_2 + S_3}$ & \textbf{78.9\%} & \textbf{63.2\%}\\
		\bottomrule
	\end{tabular}}
\end{table}
\subsection{Ablation Studies}
We further examine the impact of each subnets on the final performance on the PE challenge test set and our PE129 dataset. For simplicity, we denote the candidate proposal subnet, spatial transformation subnet and FP removal subnet as $\mathrm{S_1}$, $\mathrm{S_2}$ and $\mathrm{S_3}$ respectively. Quantitative results in terms of sensitivity at 2 FPs/scan at 0mm localization error are reported in Table \ref{tab1}. On the PE challenge test set, our proposal subnet achieves a sensitivity of 71.9\% at 2FPs. Directly removing FPs using the proposal cubes could improve the sensitivity by 1.8\%. Integrating the vessel-aligned representation using $\mathrm{S_2}$ could further improving the sensitivity by 5.2\%. For PE129 dataset, the proposal subnet achieves 42.1\% sensitivity at 2FPs, and the false positive removal subnet and spatial transformation subnet could improve the sensitivity by 2.6\%, 21.1\%, respectivity. Greater improvements achieved by $\mathrm{S_2}$ on our PE129 dataset than PE challenge might be due to many small emboli with various rotations in our dataset. Aligning them with the vessel orientation can effectively reduce the variations. 

\section{Conclusion}
This work presents an end-to-end PE detection network which achieves superior performance to the state-of-the-art methods. The main advantages of our network are three folds. First, it takes full advantage of 3D contextual information of a 3D CTPA volume via 3D FCN feature extraction. The FCN features are shared among three 3 subnets. Second, it utilizes an affine transformation layer with a ROI pooling layer to realize the vessel-aligned fixed-sized representation for arbitrary size PE candidates. The two layers can effectively reduce the appearance variations and meanwhile are completely derivable to enable end-to-end training. Third, it facilitates joint optimization of all steps for PE detection. 

%
%
%
\bibliographystyle{splncs04}
%

\end{document}